# Modeling Multiple Annotator Expertise in the Semi-Supervised Learning Scenario

**Yan Yan**[*] **Rómer Rosales**[+] **Glenn Fung**[+] **Jennifer Dy**[*]

[*]Dept. of Electrical and Computer Eng.
Northeastern University
Boston, MA

[+]IKM CAD and Knowledge Solutions
Siemens Healthcare
Malvern, PA

## Abstract

Learning algorithms normally assume that there is at most one annotation or label per data point. However, in some scenarios, such as medical diagnosis and on-line collaboration, multiple annotations may be available. In either case, obtaining labels for data points can be expensive and time-consuming (in some circumstances *ground-truth* may not exist). Semi-supervised learning approaches have shown that utilizing the unlabeled data is often beneficial in these cases. This paper presents a probabilistic semi-supervised model and algorithm that allows for learning from both unlabeled and labeled data in the presence of multiple annotators. We assume that it is known what annotator labeled which data points. The proposed approach produces annotator models that allow us to provide (1) estimates of the *true* label and (2) annotator variable expertise for both labeled and unlabeled data. We provide numerical comparisons under various scenarios and with respect to standard semi-supervised learning. Experiments showed that the presented approach provides clear advantages over multi-annotator methods that do not use the unlabeled data and over methods that do not use multi-labeler information.

## 1 Introduction

Advances in information technology have made it possible to collect data at increasingly faster rates. This has triggered and favored collaborative or aggregative forms of data collection; an example of the *Crowdsourcing* phenomenon [10]. For instance, on-line data about many specific subjects (product, topic, news article, image) is very often analyzed or processed (annotated, rated, commented on) by a multitude of individuals or entities in general. Clear instances of this include Wikipedia, most forms of multi-customer product ratings, and on-line user behavior in general.

This form of data organization creates new machine learning problems associated with the efficient utilization, modeling, and processing of such information. There are various ways to look at this problem. Translated to the supervised learning context, this problem amounts to having not one labeler (normally an expert or ground-truth) but many labelers. This novel scenario renders traditional supervised learning sub-optimal but also creates exciting new problems. The reason for this can be highlighted by noticing that now the learning algorithm can have access to a labeler (pseudo) identity in addition to the usual label values. This turns out to be a key piece of information that in many cases can have interesting implications for learning.

The multi-labeler setting is important to address real problems for which supervised learning is not suitable. These include the case when ground-truth is by nature not available (*e.g.,* what are the best results for this search query) or expensive to obtain (*e.g.,* a biopsy can provide ground-truth about cancer lesions but at a high cost).

Recently, several approaches have been undertaken to address this scenario. In particular, [15, 11] considered the case where each labeler can be modeled by associating an overall accuracy (or specificity-sensitivity values) across all the data. A more general idea that we have explored is the case where the labeling accuracy may be dependent on the actual data point observed (*e.g.,* noisy images are more difficult to label accurately than sharp images) or further dependent on annotator specific preferences [19].

This paper addresses a different facet of this problem. We draw some parallel with semi-supervised learning and explore the question of how we can exploit data that has not been annotated by any labeler or that has only been annotated by some labelers. One natural approach would consist on ignoring any unlabeled data point. While this is valid and appropriate under some models, it is not be very efficient. As will be seen, previous multi-labeler models would treat the unlabeled data in this manner. However, in this paper we propose a different strategy based on using the properties of the unlabeled data distribution.

## 2 Related work

In recent years, many semi-supervised learning methods for classification have been introduced [1, 2, 4, 5, 9, 12, 13, 20]. A complete comprehensive review on semi-supervised learning algorithms is provided in [21]. Two main scenarios have been commonly considered when training a semi-supervised model: the transductive and inductive scenarios. In the transductive setting, the learner needs to observe the unlabeled testing data while training; and therefore, although accurate, these transductive models need to be retrained (or updated) every time a test sample is to be classified. As a result, transductive algorithms may not satisfy the run-time requirements for many real-world applications, including medical diagnosis applications where new patient cases need to be classified in real-time as part of the physician's workflow. In the inductive setting, testing data is not assumed to be present at the time of model training.

Most state-of-the art approaches for semi-supervised learning are based on a weighted graph (*e.g.,* the graph Laplacian) [1, 4, 5, 13, 20] where labeled and unlabeled points constitute the vertices of a graph and the similarities between the data point pairs are represented by its edge weights. Given this graph that contains information about the spatial proximity of the training data (labeled and unlabeled), the main idea behind these methods is the notion that the classification function to be learned should give similar values for neighboring points. In other words, the value of the separator function should change smoothly over neighboring data points.

In this paper, we present an inductive semi-supervised algorithm that not only takes advantage of the available unlabeled data but also assumes that for each training point there are several labels available from different annotators (multiple labelers).

The problem of building classifiers in the presence of multiple labelers has been receiving increasing attention. One of the reasons for the increased interest in multi-labeler classification problems is that, as illustrated recently [16], employing multiple non-expert annotators can be as effective as employing one expert annotator when building a classifier. This setting is very convenient, for example, in many medical applications where the cost of expert labeling (medical specialist) is very high while non-expert annotator time is considerably less expensive (technicians). Some interesting medical application areas for multi-labeler learning include computer-aided diagnosis and radiology [15, 18] and clinical data integration [7]. However, the application areas for multiple-labeler learning can vary widely and the interest in this type of modeling is increasing rapidly. They include natural language processing, [16] computer vision [17], and product ratings or many forms of on-line collaboration.

We are not aware of previous work for solving multi-labeler classification problem in the semi-supervised scenario (combining multiply-labeled data with unlabeled data). Another distinguishing factor in this paper is that unlike any previous approaches, with the exception of [19] for supervised learning, it is not assumed that the annotator expertise is consistent across all the input data. This is a flawed assumption in many instances since annotator accuracy may depend strongly on the characteristics of a given case. Taking this into consideration, in this paper the classifiers are build so that they take into account that some labelers are better at labeling some type of data points (compared with other data points), and thus a model of the annotator expertise is also obtained.

## 3 Formulation

Let us consider $N$ data points $\{\mathbf{x}_1, \ldots, \mathbf{x}_N\}$, where $\mathbf{x}_i \in \mathbb{R}^D$. Each data point has been annotated by $T$ or fewer labelers/annotators. We denote the label provided to the $i$-th data point by annotator $t$ as $y_i^{(t)} \in \mathcal{Y}$. The labels from individual labelers are not assumed to be correct or consistent with those provided by other labelers. Let us denote the true label for each $i$-th data point (which is in general not known) as $z_i \in \mathcal{Z}$. In this paper $\mathcal{Y} \equiv \mathcal{Z}$, but this is not a requirement. For compactness, we represent the data using the matrix $X = [\mathbf{x}_1^T; \ldots; \mathbf{x}_N^T] \in \mathbb{R}^{N \times D}$ and the matrix $Y = [y_1^{(1)}, \ldots, y_1^{(T)}; \ldots; y_N^{(1)}, \ldots, y_N^{(T)}] \in \mathbb{R}^{N \times T}$, where $(\cdot)^T$ denotes matrix/vector transpose.

Consider the problem where all the data points $X$ are given, but some labels are missing for some annotators (the matrix $Y$ is incomplete). Our primary goals are to produce an estimate for the ground-truth $Z = [z_1, ..., z_N]^T$, a classifier for predicting $z$ from new instances $\mathbf{x}$, and a model for the expertise of each annotator as a function of the input $\mathbf{x}$.

### 3.1 Probabilistic Model For Multiple Annotators

Let each element of $X$, $Y$, and $Z$ be a random variable in the specified domains. In the usual supervised and semi-supervised learning scenarios, each data point has been labeled once (or unlabeled, in the semi-supervised case) and normally there is no information about the labeler identity. In the multi-labeler problem addressed in this paper, points may be labeled by zero, one, or more labelers, and in addition we are aware of who labeled what points (for example in the form of a user or annotator key that could be de-identified). The fundamental question is how to optimally exploit this multi-annotator information in the semi-supervised learning setting.

In modeling the annotator characteristics, we start by es-tablishing that a label assigned by labeler $t$ to a data point $i$, denoted by $y_i^{(t)}$, depends on the *real*, but unknown label $z_i$. In addition, we let the label also depend on the coordinates (features) of the observed point $\mathbf{x}_i$. This dependency is represented by a conditional distribution $p(y_i^{(t)}|\mathbf{x}_i, z_i)$.

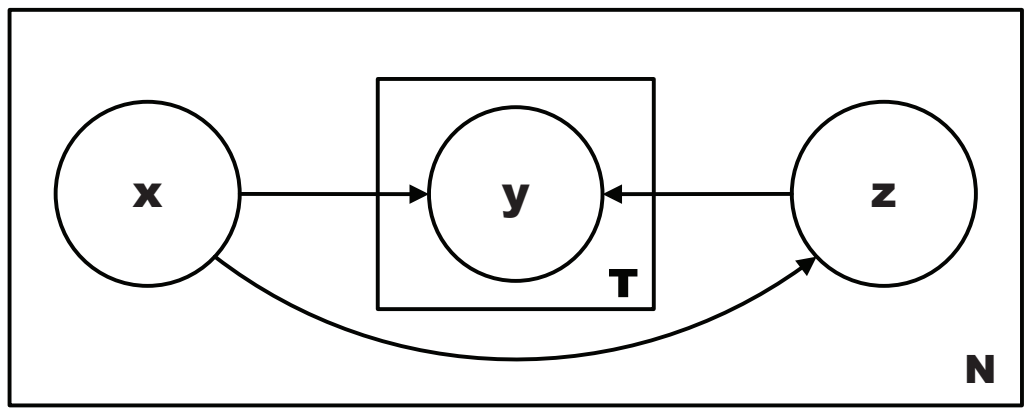

Figure 1: Bayesian network for basic multi-labeler model with variables $X$, $Y$, and $Z$.

The dependency on the true label $z_i$ alone allows us to take into consideration the differences between annotator accuracies. This dependency also suffices to model annotator's biases toward some classes and to model annotator-specific error rates for some classes. This dependency has been considered in [15, 11] to a good extent.

Now, the dependency on the input $\mathbf{x}_i$ (together with $z_i$) allows us to take into account different and interesting properties associated to the annotators. In particular, we no longer need to assume that annotators are equally good (or bad) at labeling all the data, but their accuracy depends on what input they are presented. In general, we are able to model annotator-specific and input-specific properties, such as: *for class $c$, annotator $t$ is knowledgeable (e.g., more accurate) when labeling one kind of inputs compared to other kinds of inputs*. The ability to model these properties has been recently addressed in [19]. However, this alone does not address the problem of efficiently utilizing unlabeled data as will be seen in the following section. Based on these considerations, in the following we describe a progression of three different strategies (in order of complexity) to properly incorporate unlabeled data into the model.

For our first alternative, given a data point $\mathbf{x}_i$, we posit that there is an unknown distribution $p(z_i|\mathbf{x}_i)$ that relates a point with its true label. This is basically our classification (or regression) function. Since this will imply that the labels are *independently distributed* given the observations, we call this the ID model.

The probability model just described can be represented by the graphical model depicted in Fig. 1 and can be written as (conditioned on the data points $X$):

$$p_{\mathtt{ID}}(Y, Z|X) = \prod_i p(z_i|\mathbf{x}_i) \prod_t p(y_i^{(t)}|\mathbf{x}_i, z_i). \qquad (1)$$

### 3.2 The Problem with Missing Labels

While this model is the basis of our formulation, we note that when only some data points have been annotated, with labels denoted by $Y_{\mathcal{L}}$, the model (marginal) distribution becomes:

$$p_{\mathtt{ID}}(Y_{\mathcal{L}}, Z|X) = \prod_i p(z_i|\mathbf{x}_i) \prod_{t|t\in\mathcal{T}_i} p(y_i^{(t)}|\mathbf{x}_i, z_i), \qquad (2)$$

where $\mathcal{T}_i$ is the set of annotators that provided a label for the $i$-th data point. Basically, points not labeled by any annotator will be technically ignored.

Another way to see this is by noticing that the probability of the observed labels (conditioned on the data points) does not depend on the label $z_k$ when $\mathcal{T}_k = \emptyset$:

$$\begin{aligned} p_{\mathtt{ID}}(Y_{\mathcal{L}}|X) &= [\prod_{i\backslash k}\sum_{z_i} p(z_i|\mathbf{x}_i) \prod_{t\in\mathcal{T}_i} p(y_i^{(t)}|\mathbf{x}_i, z_i)] \sum_{z_k} p(z_k|\mathbf{x}_k) \\ &= \prod_{i\backslash k}\sum_{z_i} p(z_i|\mathbf{x}_i) \prod_{t\in\mathcal{T}_i} p(y_i^{(t)}|\mathbf{x}_i, z_i) \end{aligned}$$

due to the model's conditional independence assumptions. The notation $i\backslash k$ is employed to denote set difference, in this case: $i \in \{1, 2, ..., N\} - \{k\}$.

### 3.3 Graph-Prior (GP) Alternative

In view of this and due to our interest in utilizing the (potentially large number of) unlabeled data points, we consider an alternative choice for the conditional distribution $Z|X$. This is based on incorporating a graph-based prior. For this, we consider the graph $G = (V, E)$ and associate each data point $\mathbf{x}_i$ to a node $v_i \in V$ and a weight $\phi_{ij}$ to an edge $e_{ij} \in E$. We let $z_i \in \mathbb{R}$ and in particular consider the prior given by the graph Laplacian:

$$p(Z|X) \quad \propto \quad \exp\{-\frac{Z^T \Delta Z}{2\lambda}\}, \qquad (3)$$

where $\Delta = D - \Phi$, $D = \mathtt{diag}(d_k)$, and $d_k = \sum_{j=1}^N \phi_{kj}$, for $k \in \{1, \ldots, N\}$. Thus, $\Delta \in \mathbb{R}^{N\times N}$. $\phi_{ij} \in \mathbb{R}$ is a similarity weight between data points $i$ and $j$. Intuitively, $\Phi = [\phi_{ij}]$ is a way to represent the manifold structure of the data. As an example, the Gaussian kernel can be used to define similarity weights:

$$\phi_{ij} \propto \exp\{-(\mathbf{x}_i - \mathbf{x}_j)^T \Sigma (\mathbf{x}_i - \mathbf{x}_j)\}, \qquad (4)$$

where $\Sigma$ is a positive definite matrix representing a valid distance measure. This could, for example, relate points that are *closer* to each other more heavily than those that are farther apart. The graph Laplacian has been extensively used in semi-supervised learning approaches [21]. In this paper, we borrow from this concept and adapt it to the multi-labeler scenario proposed.

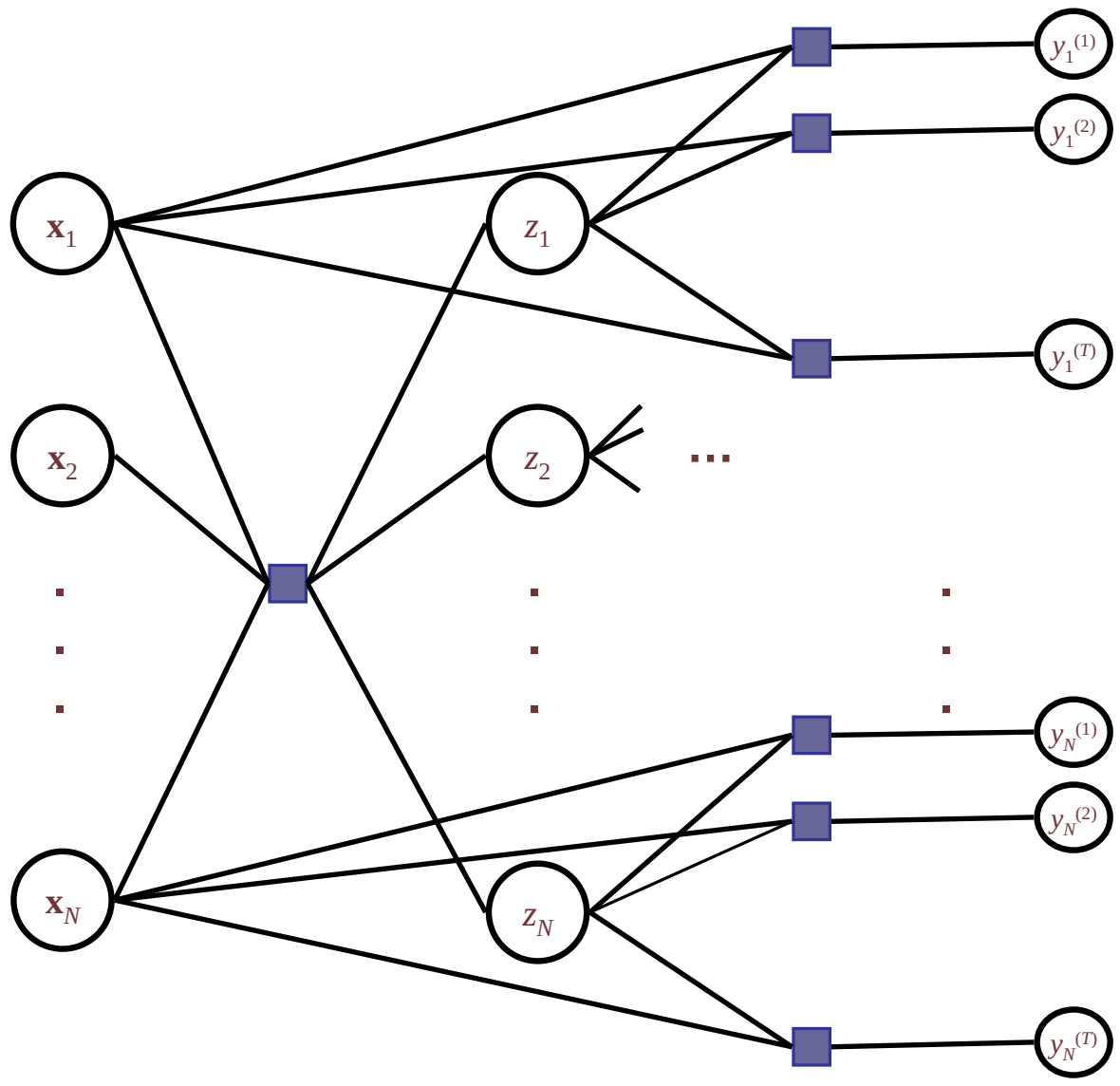

Figure 2: Factor Graph for semi-supervised multi-labeler model GP with variables $X$, $Y$, and $Z$.

Using this definition we have the model likelihood:

$$p_{\texttt{GP}}(Y_{\mathcal{L}}, Z|X; \boldsymbol{\theta}) = p(Z|X) \prod_i \prod_t p(y_i^{(t)}|\mathbf{x}_i, z_i; \boldsymbol{\theta}), \quad (5)$$

whose factor graph is shown in Fig. 2. We use $\boldsymbol{\theta}$ to denote all the model parameters, which will be considered in more detail in Sec. 3.5.

**Algorithms for Learning**

**For model** ID, using the maximum likelihood criterion, our goal is to maximize the log likelihood:

$$\log p_{\texttt{ID}}(Y_{\mathcal{L}}|X) = \sum_i \log \sum_{z_i} p(z_i|\mathbf{x}_i) \prod_{t \in \mathcal{T}_i} p(y_i^{(t)}|\mathbf{x}_i, z_i) \quad (6)$$

which does not directly lend itself to an efficient algorithm because of the log sum operation. Using Jensen's inequality and the concavity of the logarithm function we have the lower bound [6]:

$$\log p_{\texttt{ID}}(Y_{\mathcal{L}}|X) \geq \sum_{i,z_i} [\log p(z_i|\mathbf{x}_i) + \sum_{t \in \mathcal{T}_i} \log p(y_i^{(t)}|\mathbf{x}_i, z_i)],$$

which could be used as a surrogate function for maximization or as the basis for the more commonly employed Expectation Maximization (EM) algorithm, which is derived next.

There are multiple ways to develop an EM-type algorithm since both $z_i$ and $y_i^{(t)}$ could be missing (for any value of $i$ or $t$). The general recipe for the EM algorithm prescribes computing expectations for all missing random variables as part of the E-step.

The form of the likelihood (Eq. 6) makes it natural to just consider the expectations for latent variables $\{z_i\}$ and then compute exact marginals for the remaining variables (which is tractable in this case). This leads to:

*E-Step*: Compute expectations

$$\begin{aligned} \tilde{p}(z_i) &\triangleq p(z_i|Y, X) \\ &\propto p(z_i, y_i|\mathbf{x}_i) = \prod_{t \in \mathcal{T}_i} p(y_i^{(t)}|\mathbf{x}_i, z_i) p(z_i|\mathbf{x}_i) \end{aligned} \quad (7)$$

*M-Step*: Maximize $f_{\texttt{ID}}$

$$f_{\texttt{ID}} = \sum_i \sum_{t \in \mathcal{T}_i} E_{\tilde{p}(z_i)}[\log p(y_i^{(t)}|\mathbf{x}_i, z_i) + \log p(z_i|\mathbf{x}_i)]$$

**For model** GP, the distribution does not factorize in a simple manner. The equivalent bound:

$$\begin{aligned} \log p_{\texttt{GP}}(Y_{\mathcal{L}}|X) \geq \sum_Z [\log p(Z|X) + \\ \sum_i \sum_t \log p(y_i^{(t)}|\mathbf{x}_i, z_i)], \end{aligned} \quad (8)$$

is also not factorizable.

In general both E and M steps cannot be computed efficiently due to the large number of dependencies implied by the graphical model, which translates into a summation over all possible combinations of values for $Z$ (this grows exponentially with the number of data points).

Practical alternatives for learning this model are possible, in particular to reduce the complexity of estimating $\tilde{p}(z_i)$ (using approximations). For example, one could consider the $m$ neighbors of each data point, or more generally those that have the largest (direct) influence on the calculation and approximate the posterior (required in the E-step of EM) as $p(z_i|X) \approx p(z_i|\mathbf{x}_i, \mathbf{x}_{\eta(i)})$, where $\eta(i)$ is the index set for the neighbors of the $i$-th data point $\mathbf{x}_i$ using a metric of choice. While this is possible in practice, our numerical experiments suggest that the following section provides a better alternative.

### 3.4 A Logistic + Graph-Prior Model (LGP)

This variation on the graph-prior model of Sec. 3.3 addresses two potential issues: (1) the posterior $p(z_i|X)$ required an approximation; this can be a limitation (we are not aware of any practical approximation guarantees for the true posterior). A more important limitation is given by the fact that (2) the prior distribution $p(Z|X)$ (Eq. 3) is technically fixed beforehand (albeit the scaling parameter $\lambda$ could be adjusted; for example, via cross-validation); thus, limiting the model flexibility.

In this model, let us introduce a new parameter $\xi \in \mathbb{R}^D$ that will allow us to provide a more flexible prior. First,

consider the following logistic model for the true label $z_i$:

$$p(z_i = 1|\xi, \mathbf{x}_i) = (1 + \exp(-\xi^T \mathbf{x}_i))^{-1}. \quad (9)$$

This has the advantage that $z_i$ depends only on $\mathbf{x}_i$; however as we have seen, this assumption will not allow our model to take advantage of all the unlabeled data. This situation can be remedied by placing a graph prior for the *new* parameter $\xi$:

$$p(\boldsymbol{\xi}|X) \propto \exp\{-\boldsymbol{\xi}^T X^T \Delta X \boldsymbol{\xi}\} = \exp\{-\boldsymbol{\xi}^T A \boldsymbol{\xi}\}, \quad (10)$$

where $A \triangleq X^T \Delta X$ and $\Delta$ is the graph Laplacian defined in Sec. 3.3. Combining these definitions we have

$$p(Z, \boldsymbol{\xi}|X) = \prod_{i=1}^{N} p(z_i|\boldsymbol{\xi}, \mathbf{x}_i) p(\boldsymbol{\xi}|X), \quad (11)$$

leading us to a new model that can be written as:

$$\begin{aligned} p_{\texttt{LGP}}(Y_{\mathcal{L}}, \xi|X) &= \sum_Z p(Z, \xi|X) p(Y_{\mathcal{L}}|Z, X) \\ &= p(\boldsymbol{\xi}|X) \times \\ &\quad \prod_i \sum_{z_i} [p(z_i|\xi, \mathbf{x}_i) \prod_{t \in \mathcal{T}_i} p(y_i^{(t)}|\mathbf{x}_i, z_i)]. \end{aligned}$$

### Algorithms for Learning

**For** LGP (Logistic-GP), using the maximum likelihood criterion and the EM algorithm we have:

*E-step*: Compute expectations

$$\begin{aligned} \tilde{p}(z_i) &\triangleq p(z_i|X, Y_{\mathcal{L}}, \xi; \boldsymbol{\theta}) \quad (12) \\ &\propto p(\xi|X) p(z_i|\mathbf{x}_i, \xi) \prod_{t \in \mathcal{T}_i} p(y_i^{(t)}|\mathbf{x}_i, z_i; \boldsymbol{\theta}). \end{aligned}$$

*M-step*: Maximize $f_{\texttt{LGP}}$

$$\begin{aligned} f_{\texttt{LGP}} &= E_{\tilde{p}(\mathbf{z})}[\log p(\mathbf{z}, Y_{\mathcal{L}}|X, \xi; \boldsymbol{\theta})] \\ &= \sum_i \sum_{t \in \mathcal{T}_i} E_{\tilde{p}(z_i)}[\log p(y_i^{(t)}|\mathbf{x}_i, z_i; \boldsymbol{\theta}) \\ &\quad + \log p(z_i|\xi, \mathbf{x}_i) + \log p(\xi|X)], \quad (13) \end{aligned}$$

where the graph prior $p(\xi|X)$ depends on the data, but remains the same once $X$ has been observed. The overall MAP estimate for $\xi$ conditioned on all the observed variables is updated iteratively as shown in Algorithm 1. Thus, for the M-step we optimize $f_{\texttt{LGP}}$ with respect to $\theta$ and $\xi$.

### 3.5 Specific Choice of Distributions

While the model structures have been defined in the previous sections, we have yet to instantiate the specific form of the various distributions employed. The fundamental multi-labeler conditional distribution in this paper is defined as follows:

$$p(y_i^{(t)}|\mathbf{x}_i, z_i) = \mathcal{N}(y_i^{(t)}; z_i, \sigma_t(\mathbf{x}_i)), \quad (14)$$

where the variance depends on the input $\mathbf{x}$ and is also specific to each annotator $t$.

In this paper we have $y^{(t)} \in \{0, 1\}$ and thus we found appropriate to let $\sigma_t(\mathbf{x}) \in (0, 1]$ by setting $\sigma_t(\mathbf{x})$ as a logistic function of $\mathbf{x}_i$:

$$\sigma_t(\mathbf{x}) = (1 + \exp(-\mathbf{w}_t^T \mathbf{x}_i - \gamma_t))^{-1}. \quad (15)$$

An alternative model such as a Bernoulli distribution (replacing Eq.14) could be employed if we needed to restrict $y^{(t)}$ to be binary.

In order to perform the M-step, we use the L-BFGS quasi-Newton method requiring only to calculate the gradient direction. The partial derivatives with respect to the various parameters are given by:

$$\frac{\partial f_*}{\partial \mathbf{w}_t} = \sum_{i=1}^{N} E_{\tilde{p}(z_i)}[\tfrac{(y_i^{(j)} - z_i)^2}{\sigma_t^2(\mathbf{x}_i)} - 1](1 - \sigma_t(\mathbf{x}_i))\mathbf{x}_i \quad (16)$$

$$\frac{\partial f_*}{\partial \gamma_t} = \sum_{i=1}^{N} E_{\tilde{p}(z_i)}[\tfrac{(y_i^{(j)} - z_i)^2}{\sigma_t^2(\mathbf{x}_i)} - 1](1 - \sigma_t(\mathbf{x}_i)) \quad (17)$$

For ID, we let $p(z_i|x_i)$ be the logistic model:

$$p(z_i = 1|\mathbf{x}_i) = (1 + \exp(-\alpha^T \mathbf{x}_i - \beta))^{-1}, \quad (18)$$

where the parameters $\alpha$ and $\beta$ are obtained also during the M-step. The gradients are given by:

$$\begin{aligned} \frac{\partial f_*}{\partial \alpha} &\propto \sum_i \frac{\delta_{\tilde{p}} \exp(-\alpha^T \mathbf{x} - \beta)\mathbf{x}}{(1 + \exp(-\alpha^T \mathbf{x} - \beta))^2} \\ \frac{\partial f_*}{\partial \beta} &\propto \sum_i \frac{\delta_{\tilde{p}} \exp(-\alpha^T \mathbf{x} - \beta)}{(1 + \exp(-\alpha^T \mathbf{x} - \beta))^2}, \end{aligned}$$

where $\delta_{\tilde{p}} = \tilde{p}(z_i = 1) - \tilde{p}(z_i = 0)$.

For LGP (Logistic-GP), we require a few additional calculations. The new gradient is given by:

$$\begin{aligned} \frac{\partial f_*}{\partial \xi} &= \frac{1}{p(\xi|X)} \cdot \frac{\partial p(\xi|X)}{\partial \xi} \\ &\quad \sum_i \frac{\sum_{z_i} [\partial p(z_i|\xi, \mathbf{x}_i)/\partial \xi] \prod_t \alpha(y_i^{(t)}; \theta_t)}{\sum_{z_i} p(z_i|\xi, \mathbf{x}_i) \prod_t \alpha(y_i^{(t)}; \theta_t)} \end{aligned}$$

$$\frac{\partial f_*}{\partial \theta_s} = \sum_i \frac{\sum_{z_i} p(z_i|\boldsymbol{\xi}, \mathbf{x}_i) \prod_{t \backslash s} \alpha(y_i^{(t)}; \theta_t) \frac{\partial \alpha(y_i^{(s)}; \theta_s)}{\partial \theta_s}}{\sum_{z_i} p(z_i|\boldsymbol{\xi}, \mathbf{x}_i) \prod_t \alpha(y_i^{(t)}; \theta_t)},$$

where we have used $\alpha(y_i^{(t)};\theta_t) = p(y_i^{(t)}|\mathbf{x}_i, z_i;\theta_t)$ to simplify the notation. The parameter $\theta$ is composed of $\{\mathbf{w}_t\}_{t=1}^T$ and $\{\gamma_t\}_{t=1}^T$.

The general learning approach can be summarized in Algorithm 1. The same algorithm can be used for any of the discussed model by replacing the appropriate gradients.

**Algorithm 1** Multi-Annotator Semi-Supervised Learning

input: $X, Y$; set: $\mathbf{w}_t = \mathbf{0}$, $\gamma_t = 0$ for $t = 1, \ldots, T$ and threshold $\epsilon$
initialize: $\mathbf{w}_{t\texttt{new}}, \gamma_{t\texttt{new}}$
**while** $\sum_{t=1}^{T} [\|\mathbf{w}_t - \mathbf{w}_{t\texttt{new}}\|^2 + (\gamma_t - \gamma_{t\texttt{new}})^2] \geq \epsilon$ **do**
  E-step: estimate $\tilde{p}(z_i)$ for every $\mathbf{x}_i$ in data $X$
  M-step:
  1)Update $\mathbf{w}_{t\texttt{new}}$, $\gamma_{t\texttt{new}}$ that maximize $E_{\tilde{p}(\mathbf{z})}[\log p(\mathbf{z}, Y|X, \boldsymbol{\theta})]$ using the LBFGS quasi-Newton approximation to compute the step, with gradient equations (16,17).
  2)Update the estimates for $\alpha, \beta$ (for `ID`) or $\xi$ (for `LGP`) using the appropriate gradients.
**end while**
**return** $\{\tilde{p}(z_i)\}, \{\mathbf{w}_t\}, \{\gamma_t\}; t = 1, \ldots, T, i = 1, \ldots, N$

### 3.6 Prediction

Given a learned model, there exist multiple ways to interpret the problem of making a label prediction given a new data point. Here we focus on inferring the ground-truth for a new data point that was not known during training time (the usual inductive scenario). Specifically, we would like to estimate $p(z_{\texttt{test}}|X, Y_{\mathcal{L}};\boldsymbol{\theta})$. This is basically equivalent to performing the E-step as described before and computing $\tilde{p}(z_{\texttt{test}})$ as in the model-specific E-step.

## 4 Experiments

In this section, we compare our final semi-supervised multi-annotator model, the logistic graph Laplacian prior (Logistic-GP), against baseline methods on a number of UCI Machine Learning Repository [8] benchmark data with simulated labelers, and a real data set with multiple labelers for the problem of automatic assessment of heart wall motion abnormalities (AWMA) with information extracted from ultrasound images [14]. In our experiments we show the results for our logistic + graph prior (`LGP`). Compared to the `ID` model, this has the clear advantage of being able to make use of the unlabeled data as shown in Sec. 3.2.

Since there are no existing semi-supervised multi-annotator models in the literature, we compare our method against the following baselines, testing different aspects of our model: (1) standard logistic regression classifier trained on labels from the annotators' majority vote (we call *majority vote*), (2) standard logistic regression classifiers trained on labels from each annotator (we call *annotator* $t$), (3) a supervised multi-labeler logistic regression model version of our approach with the variance not a function of the input $x$ (*ML original*), which is similar in spirit to that of [11, 15], and (4) a semi-supervised support vector machine (SVM) classifier with a linear kernel from SVM-light[1] (*SVM-light*) trained on labels from the annotators' majority vote. The parameters in SVM were tuned on a validation set using grid search. We compare against methods 1 to 3 to test the advantage of learning from unlabeled data. In addition, by comparing with 1, 2 and 4, we also test whether or not learning from multi-labelers is better than just from one labeler alone. Against method 3, we also test the effect of taking the variance of an annotator's accuracy ($\sigma(x)$) in labeling across different observations into account to classification performance.

### 4.1 UCI Benchmark Data

We performed experiments on various datasets from the UCI machine learning data repository [8]: Ionosphere $(351, 34)$, Dim $(4192,14)$, Housing $(506, 13)$, Pima $(768,8)$, BUPA $(345,6)$, Wisconsin Breast Cancer 24 $(155,32)$, and Wisconsin Breast Cancer 60 $(110,32)$, where the numbers in parenthesis indicate the number of samples and features (data dimensionality) respectively. Since multiple annotations for any of these UCI datasets are not available, we need to simulate several labelers with different *labeler expertise* or accuracy. In order to simulate the labelers, for each dataset, we proceeded as follows: first, we clustered the data into five subsets using k-means [3]. Then, we assume that each one of the five simulated labelers $i, i = 1 \ldots 5$ is an expert on cases belonging to cluster $i$, where their labeling coincides with the ground truth; for the rest of the cases (cases belonging to the other four clusters), labeler $i$ makes a mistake $35\%$ of the time (we randomly switch labels for $35\%$ of the data samples). Figure 3 displays plots of the stratified five-fold cross-validated accuracies of the different methods on these UCI datasets as the proportion of the training data that is labeled is increased. These results show that our semi-supervised multi-labeler based on the logistic and Laplacian prior (Logistic-GP) has the best accuracies in almost all proportion of labeled training data cases for all the datasets. Note that our approach performed better than SVM-Light because we were able to take into account multiple labelers expertise into our model. It is better than the supervised methods because we are able to learn from unlabeled data as well. Moreover, it is better than ML original because our method is semi-supervised and also because we take the effect of the variance of an annotator's accuracy ($\sigma(x)$) in labeling across different observations into account.

[1]http://svmlight.joachims.org/

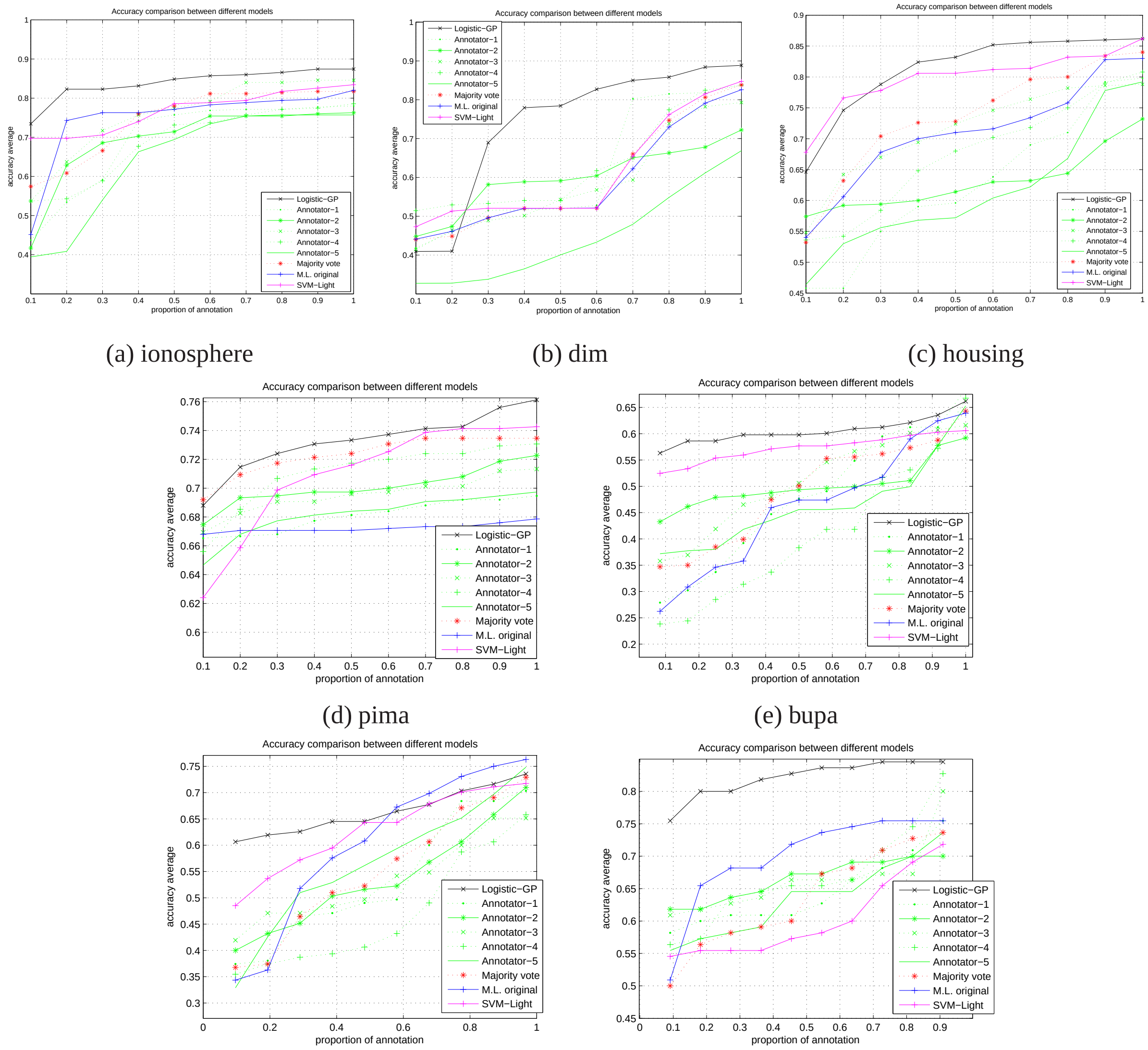

(a) ionosphere (b) dim (c) housing

(d) pima (e) bupa

(f) breast cancer 24 (g) breast cancer 60

Figure 3: Accuracies for the various UCI datasets for different proportion of labellings in the training data (0.1 - 1.0). Results show averages for five randomized splits of training (labeled and unlabeled) and test sets, with cross-validation

## 4.2 AWMA Cardiac Data

The Automatic Wall Motion Abnormality Detection (AWMA) data consists of 220 ultrasound image sequences of the heart motion (generated using pharmacological stress). All the cases have been labeled at the heart wall segment level by a group of five trained cardiologists. According to standard protocol, there are 16 left ventricle heart wall segments. Each of the segments were ranked from 1 to 5 according to its movement. For simplicity, we converted the labels to a binary (1 = normal, 2 to 5 = abnormal). This data provides us with sixteen two-class classification problems (one problem for each segment). For our experiments, we used 24 global and local image features for each node calculated from tracked contours. Since we have five doctor labels but no ground we will assume that the majority vote of the five doctors are a fair approximation to the true labels. We applied stratified five-fold cross-validation to evaluate our results. Figure 4 shows the average five-fold cross-validated accuracies of our method against the different baselines as we increase the proportion of training data labeled by annotators. Since this data is actually comprised of several classification problems (one for each segment), we report the average (top) and the standard deviation results (bottom). We observe that our model outperforms all the baseline methods for this data in terms of average accuracies. This is because we have extra unlabeled information helping us improve our performance and we take multiple annotator labels into account in learning our classifier.

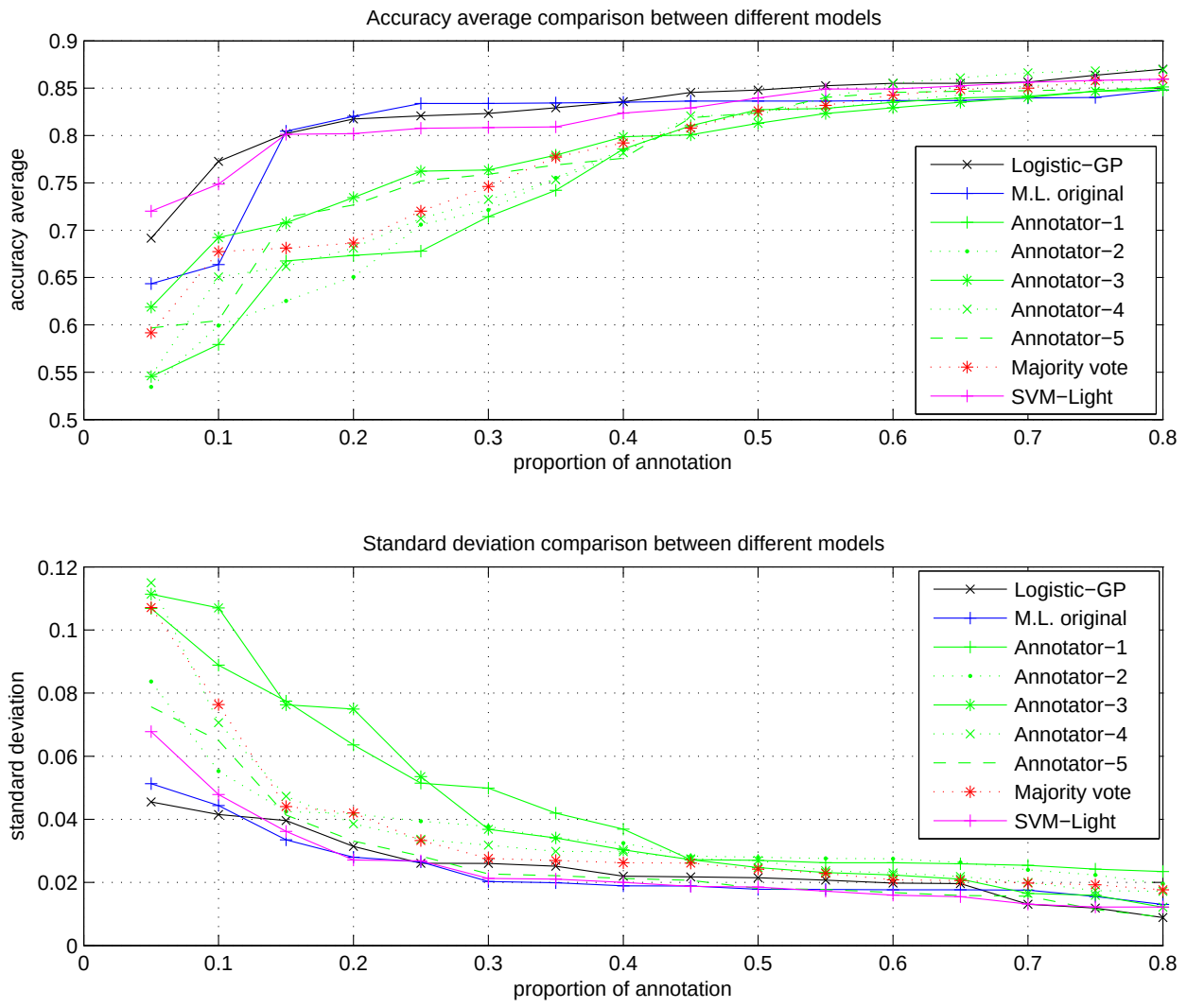

Figure 4: Accuracies for the AWMA cardiac data with different percentage of the training data labeled, averaged over five randomized train/test sets selection

Accuracies do not reveal the trade-off between the true positive rate and false alarm rate. Here, we also show the average ROC results for our model compared to the different baselines. We set the proportion of training points to 20%, 50% and 100%. The results are shown in Figure 5. The Logistic-GP (LGP) model outperform the others when there are only few labeled points (at 20% of the training data). Logistic-GP is slightly better than semi-supervised SVM (SVM-light) and almost equal in performance to ML original when using all 100% labeled (i.e. no extra unlabeled information). SVM-light performed reasonably well and close to our method when there are a few labels because the ground truth for this data is based on the majority vote of the annotators which is what SVM-light uses as labels for training. Interestingly, even though Logistic-GP does not have this ground truth label, but labels from all five labelers, it outperformed SVM-light. Note, that when all labels are available, Logistic-GP together with ML original, both of which learn from multiple labelers, outperformed the others which are simply based on the majority label (ground-truth) or labels from single annotators.

## 5 Conclusions

Most classification algorithms are designed to utilize labels as if they were provided by just one annotator. However, in some applications multiple labels for the same data point, provided by multiple annotators, may be available. This paper addresses how to learn from all the different annotations to build a better classifier compared to classification algorithms that can only learn from one labeler. From our experience working in applications where multiple annotators are available (*e.g.*, medical experts), we observed that typically the different annotators have varying expertise, and importantly this expertise varies based on the observation being labeled. We have thus incorporated this variability in labeling as a function of the observation ($\mathbf{x}$) in our model. In addition to learning from multiple annotator information, we also allow our model to learn from unlabeled data. This ability is important in many common domains where there exist large amounts of unlabeled data.

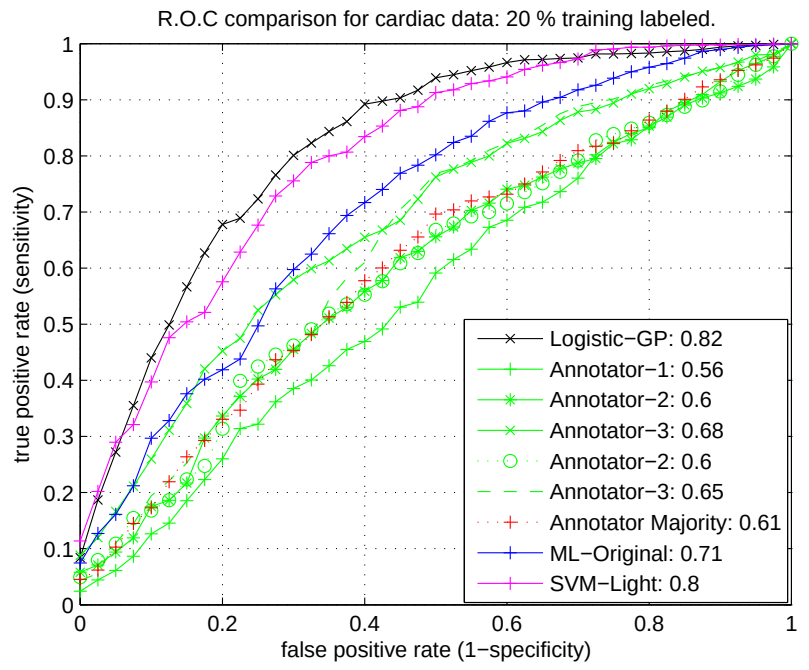

(a) 20%

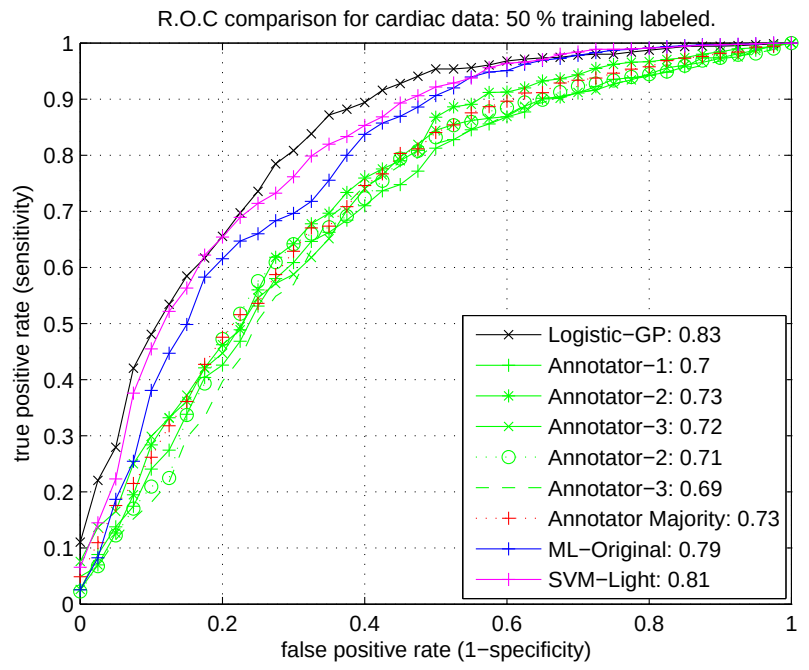

(b) 50%

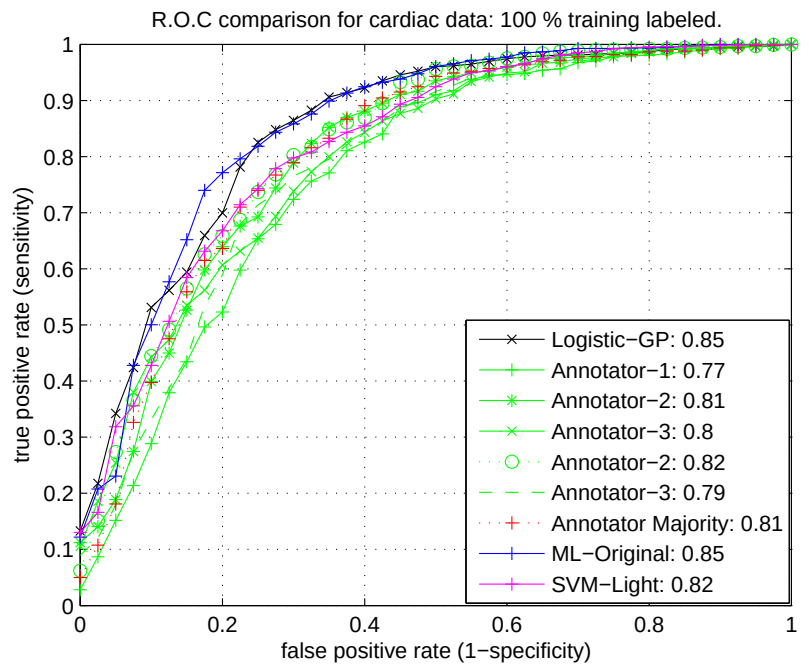

(c) 100%

Figure 5: ROC comparison with (a) 20%, (b) 50% and (c) 100% of the training data labeled for the AWMA data.

In summary, in this paper we introduced a probabilistic model that can properly learn from data for which several labels, one, or no labels are available for each data point. In addition, we model annotators' varying expertise across the input space. Our experiments employing real and simulated annotators on UCI benchmark and real medical data show that our model, taking advantage of the extra information from the unlabeled data, outperforms approaches that only learn from labeled data. Moreover, learning from multiple annotators improves classification performance over standard single annotator supervised and semi-supervised classifiers.

**Acknowledgements**

This work is supported in part by NSF IIS-0915910.